\documentclass[runningheads]{llncs}

\usepackage{amsmath} 
\usepackage{graphicx}
\usepackage{makecell}
\usepackage{pdflscape} 
\usepackage{multirow}  
\usepackage{float} 
\usepackage{longtable}
\usepackage{rotating}
\usepackage{xcolor}
\usepackage{hyperref}
\usepackage[markup=underlined]{changes} 
\definechangesauthor[color=blue]{anon}
\usepackage{placeins} 
\usepackage{subcaption}
\usepackage[ruled,linesnumbered]{algorithm2e}
\usepackage{times}
\usepackage{relsize}

\newif\ifworkinprogress
\workinprogresstrue

\ifworkinprogress
	\newcommand{\sa}[1]{\textcolor{magenta}{{[Simone] #1}}}

\else
  \newcommand{\sa}[1]{}
  \newcommand{\gh}[1]{}
  \newcommand{\asa}[1]{}
  \newcommand{\am}[1]{}
\fi

\sloppypar

\newcommand{\MA}{\mathcal{A}}

\newcommand{\MB}{\mathcal{B}}
\newcommand{\avg}{\operatorname{avg}}

\begin{document}
\title{Accurate and Noise-Tolerant Extraction of Routine Logs in Robotic Process Automation \\ (Extended Version)}
\titlerunning{Accurate and Noise-Tolerant Extraction of Routine Logs in RPA}

\author{%
  Massimiliano de Leoni\inst{1} \and
  Faizan Ahmed Khan\inst{1} \and
  Simone Agostinelli\inst{2}
}

\institute{%
  Department of Mathematics, University of Padua, Italy\\
  \email{massimiliano.deleoni@unipd.it, faizanahmed.khan@phd.unipd.it}
  \and
  Department of Engineering and Science, Universitas Mercatorum of Rome, Italy\\
  \email{simone.agostinelli@unimercatorum.it}
}

\maketitle              
\begin{abstract}
Robotic Process Mining focuses on the identification of the
routine types performed by human resources through a User Interface.
The ultimate goal is to discover routine-type models to enable robotic
process automation. The discovery of routine-type models requires the provision of a routine log. Unfortunately, the vast majority of existing
works do not directly focus on enabling the model discovery, limiting
themselves to extracting the set of actions that are part of the routines. They were also not evaluated in scenarios characterized by inconsistent routine execution, hereafter referred to as noise, which reflects natural variability and occasional errors in human performance. 
This paper presents a clustering-based technique that aims to extract routine logs. Experiments were conducted on nine UI logs from
the literature with different levels of injected noise. Our technique was
compared with existing techniques, most of which are not meant to discover routine logs but were adapted for the purpose. The results were
evaluated through standard state-of-the-art metrics, showing that we can
extract more accurate routine logs than what the state of
the art could, especially in the presence of noise.
\end{abstract}

\section{Introduction}
\label{sec:intro}
Robotic Process Automation (RPA) is a maturing technology~\cite{van2018robotic} that creates software (SW) robots to partially or fully automate rule-based and repetitive \textit{routines}
performed by human users in their applications' user interfaces (UIs)~\cite{Plattfaut.2022RPALitRev}.

Jimenez et al.~\cite{jimenez2019method} put forward a methodology to automate the routines and replace them with SW robots, which consists of three steps: \textit{(i)} identify the candidate routine types to automate by means of interviews and observation of workers conducting their daily work, \textit{(ii)} record the interactions that take place during routines’ enactment on the UI of software applications into dedicated UI logs, and \textit{(iii)}  specify their conceptual and technical structure for defining the behavior of SW robots in the form of routine-type models. Once these models have been defined, commercial RPA tools allow SW robots to automate a wide range of routines in a record-and-replay fashion 

The third step of the methodology, namely the definition of the routine-type model, can be easily achieved through the application of process discovery techniques~\cite{vanderAalst2016}. However, traditional process discovery techniques require the event log to record a set of executions of one process type. This is achieved by associating a case identifier to each event, which enables grouping events into traces, each of which represents one execution of the process that is aimed to be discovered. UI logs cannot be directly employed by traditional process discovery techniques because they do not follow the assumptions mentioned above: firstly, UI-log events do not have case identifiers that allow grouping events into individual routine executions; secondly, UI logs contain executions of different routine types, instead of one single routine type. 

It follows that the automatic discovery of routine-type models requires transforming UI logs into a set of routine logs, one per routine type. Subsequently, the events - in fact, UI events - of the routine log related to a certain type $T$ must be assigned identifiers that enable grouping the events into individual executions of $T$. In the remainder, we refer to UI events of routine logs as actions, to differentiate them from the events of traditional event logs.  Our proposal shares commonalities with techniques for event-log trace clustering, such as~\cite{zandkarimi2020generic}. However, they require event logs with case identifiers as input, which differs from the structure of UI logs. While this limitation might possibly be lifted, their applicability to UI logs and RPA has so far remained an unexplored area. 

The literature largely proposes techniques that can allocate UI actions to routine types, without attempting to create routine logs~\cite{DBLP:conf/icpm/LenoADRMP20},~\cite{DBLP:conf/icsoc/AgostinelliLM21}, ~\cite{bosco2019discovering}, with the notable exception of Rebmann et al.~\cite{DBLP:conf/caise/RebmannA23} and Abb et al.~\cite{abb2022trace}. Unfortunately, identifying the actions of different routine types is unsatisfactory, because the creation of SW robots requires modeling how these actions must be ordered. 
We thus attempted to adapt existing techniques to discover routine logs, but the results were not as satisfactory as we aimed at.
 Our goal would also have been to compare with Abb et al.~\cite{abb2022trace}, which enables direct extraction of routine logs. Unfortunately, the lack of a public reference implementation prevented us from carrying out the comparison.

Furthermore, these techniques were designed and assessed in scenarios where humans always work consistently, and thus all routine executions of a certain type were somewhat always executed in the same manner. This scenario is not realistic because humans naturally perform different executions of the same routine type in a slightly different manner and, additionally, they can make mistakes, which requires corrective actions, such as undoing mistakes and redoing the right actions. For instance, humans may copy and paste the wrong cells of an Excel file, possibly realizing their mistakes and trying to fix them through additional actions. This lack of consistency and outlier behavior is hereafter referred to as noise. If the techniques to extract routine logs are not robust against noise, routine logs are mistakenly created. As a consequence, when applying discovery techniques onto these erroneous UI logs, the resulting routine-type models may be inaccurate and imprecise.

This paper proposes a new technique that is able to extract routine logs from UI logs and is noise-tolerant. The validity of the proposal was evaluated on nine different UI logs with varying levels of injected noise, and compared against state-of-the-art techniques, which were  adapted for the purpose but, unfortunately, without achieving the same result quality as that of our proposal. 
The assessment of the accuracy in extracting routine logs has been based on the two different state-of-the-art criteria: \textit{Jaccard Coefficient (JC)} and \textit{fitness}.

The rest of the paper is structured as follows. Section \ref{sec:preliminaries} introduces the background required to understand the entire paper.
Section \ref{sec:routine-identification} presents the three-step technique to the discovery of
routine logs, while Section \ref{sec:experiments} compares it with existing techniques against standard metrics. Finally, Section \ref{sec:conclusion} concludes the paper.

\section{Preliminary Concepts}
\label{sec:preliminaries}

Here we introduce the basic concepts related to Robotic Process Mining (RPM). We aim to discover the types of routines that users perform. Each routine execution is composed of a sequence of UI actions (e.g., copying and pasting cells, clicking buttons). In this paper, we aim to discover the models of routine types from a UI log, which records the actions performed by users through the UI: 
\begin{definition}[UI Log]
\label{definition:def1}
Let $\MA$ be the set of UI actions of interest.
 A UI Log $\Sigma$ is a sequence of actions, namely $\Sigma \in \MA^*$.
\end{definition}
Note that UI logs can generally be richer than simply containing the names of the activities; for instance, the UI log could record the timestamps, the user names, etc. However, these concepts are abstracted here because they are irrelevant for the remainder of the paper. 

 We denote the length of the sequence $\Sigma$ by $|\Sigma|$, i.e., the number of actions it contains. For each index $i$ with $1 \leq i \leq |\Sigma|$, $\Sigma(i)$ denotes the $i$-th action in the sequence $\Sigma$.

The identification of the routine types requires the identification of routine executions from UI logs and their subsequent clustering: every cluster contains the execution of routines of the same type. 
Each \textbf{routine execution} $\lambda$ consists of actions performed by one single user, which is recorded as a sequence of execution of actions of a routine, namely $\lambda \in \MA^*$. 
The clustering of the routine executions is defined as follows:\footnote{Given a set $X$, $\MB(X)$ indicates the set of all multisets with elements in $X$. Given two multisets $Y$ and $Z$, $Y \uplus Z$ indicates the union of $Y$ and $Z$, namely the multiset of the elements in $Y$ combined with those in $Z$, each with a cardinality that is the sum of its cardinality in $Y$ and in $Z$.}

\begin{definition}[The Problem of Clustering Routine Executions]
\label{def:research-question}
Let $\MA$ be the set of UI actions of interest. Let $\Sigma \in \MA^*$ be a UI Log.
The problem of clustering routine executions from $\Sigma$ consists in:
\begin{enumerate}
    \item Extracting the multiset $W \in \MB(\MA^*)$ of routine executions from $\Sigma$,
    \item Clustering $W$ in a set  $\{C_1, \ldots, C_n\} \subset \MB(\MA^*)$  such that $W=\uplus_{i=1}^{n} C_i$, where $n$ is the number of routine types and $C_i$ is the routine log related to the $i$-th routine type. 
   
\end{enumerate}
\end{definition}
The extraction of routine executions is often referred to in the literature as \textbf{segmentation}~\cite{DBLP:series/lnbip/Agostinelli24}.
Each identified cluster at point 2 is a \textbf{routine log}, which contains the executions of routines in the UI log that are of the same type.
Note the use of multisets for the identified routines and routine logs. Since routines are abstracted out as actions, the same routine is likely going to occur multiple times. In fact, each routine type is expected to be independently automated through RPA Tools (e.g., the automation of the reimbursement procedure or student enrollment). 

For each routine log $C=\{\sigma_1,\ldots,\sigma_n\}$ associated with some routine type $T$, a model of $T$ can be discovered, using existing process-discovery techniques on $C$~\cite{vanderAalst2016}.
\begin{figure}[t!]
\centering
\includegraphics[width=\columnwidth]{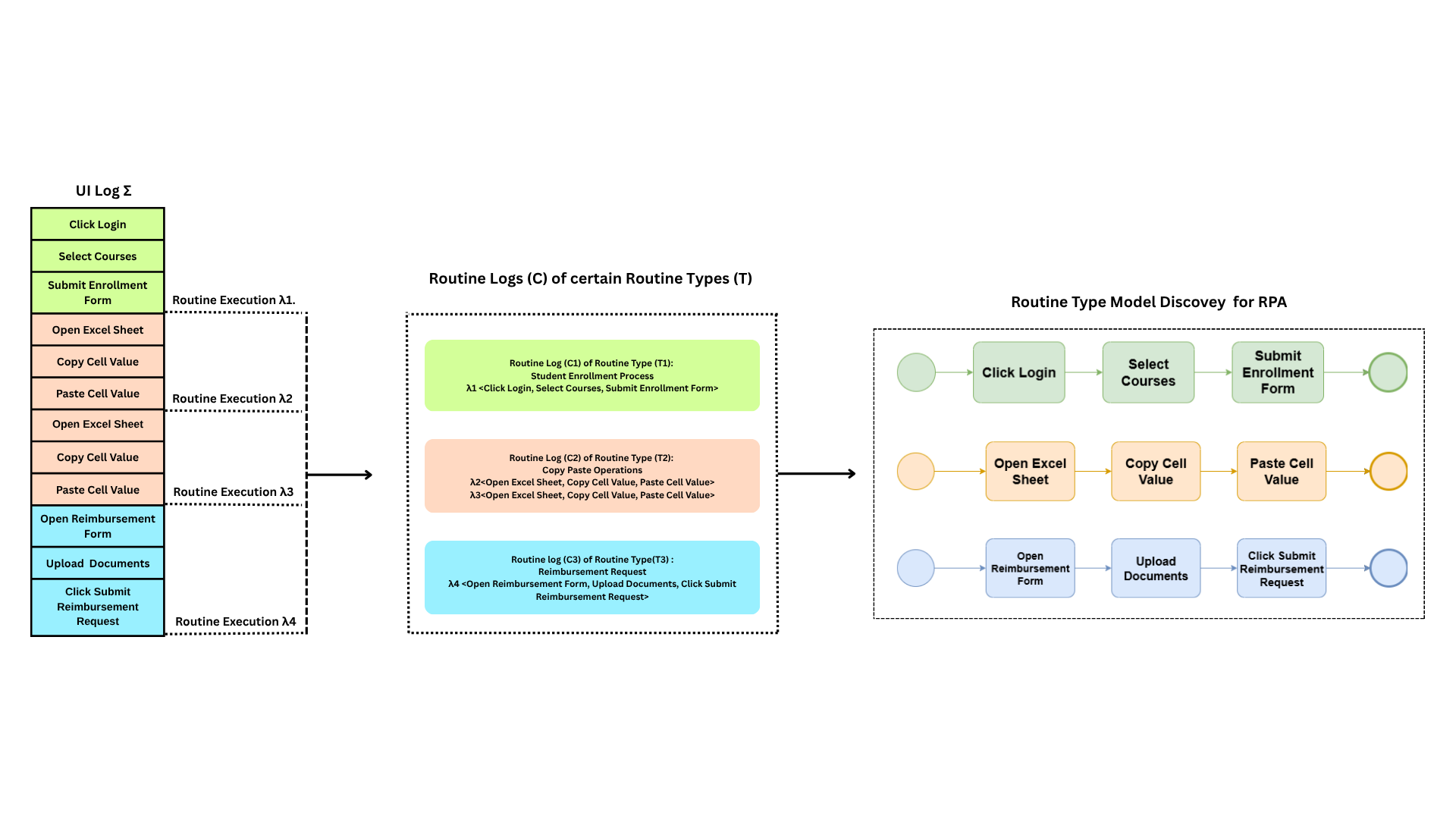}
\caption{Example illustrating the preliminary concepts: transformation from an UI log to multiple routine logs of a specific routine type.} 
\label{fig:UI_Log}
\end{figure}
The goal of this paper is to discover routine logs by addressing the problem outlined in Definition~\ref{def:research-question}. \textit{These routine logs are naturally used to discover models of routine types, but this paper does not provide a contribution in this respect. However, the assessment is also based on the fitness metric, which enables determining whether an extracted routine log is able to be used as input to discover an accordant routine type} (cf.\ Definition~\ref{def:max_fitness} in Section~\ref{subsec1:experiment}). 

Figure~\ref{fig:UI_Log} provides an example to support the understanding of these preliminary concepts. Suppose we are given the UI log shown on the left side of the figure, which records three routine executions—each belonging to a different routine type. In the figure, the three executions are visually distinguished using different colors. Naturally, in practice, neither the number of routine executions nor their respective routine types is known beforehand. The core challenge, as stated in Definition~\ref{def:research-question}, is to determine that there are indeed three routine executions and that they correspond to three distinct routine logs. Finally, as depicted on the right side of Figure~\ref{fig:UI_Log}, each identified routine log is then fed into a process discovery algorithm to derive a separate model for each routine type.

In this paper, UI logs are modeled without considering the resources that execute the routines. This simplification is not restrictive. If UI log events were defined as pairs of actions and resources, one could easily derive a sub-log for each resource $r$ by retaining only the actions performed by $r$. By subsequently concatenating the sub-UI logs of all resources and projecting solely on the actions, one would obtain a UI log that aligns with Definition~\ref{definition:def1}. Note that each execution of any routine type is naturally carried out by the same resource. Moreover, timestamps are irrelevant to the proposed technique, as will become clear in the remainder of this paper.

\section{A Technique for Discovering Routine Logs}
\label{sec:routine-identification}
\begin{figure}[t!]
\centering
\includegraphics[width=\columnwidth]{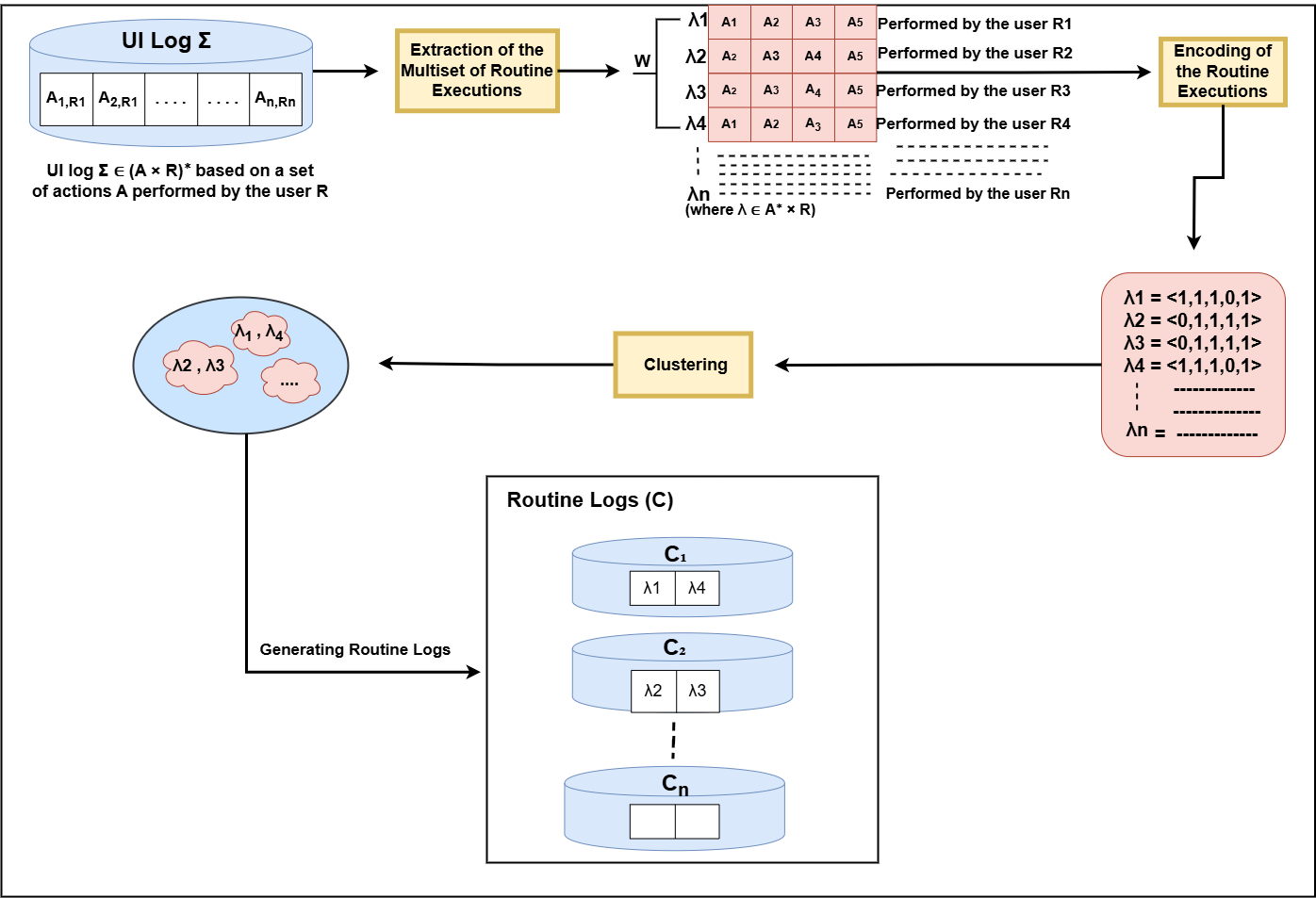}
\caption{Overview of the proposed technique for routine log discovery.} 
\label{fig:abstract}
\end{figure}

The starting point of the technique is a UI log $\Sigma \in \MA^*$ (cf.\ Definition~\ref{definition:def1}). The UI log $\Sigma $ contains different routine executions $\lambda \in \MA^* $.\ The proposed work aims to discover routine logs (namely, a multiset of routine executions) that can later be used to discover routine-type models. 
Figure \ref{fig:abstract} provides an overview of the technique: starting from $\Sigma$, the technique first extracts the multiset of routine executions, which are then encoded into respective vectors. Subsequently, these vectors are split into cluster \( C_i \). Finally, for each cluster \( C_i \), the routine executions corresponding to the vectors in each \( C_i \) are collected together to form the routine log for that cluster. 

\paragraph{Extraction of the Multiset of Routine Executions.}
The extraction of routine executions requires one to define the set $\MA_F \subset \MA$ of actions that mark the completion of a routine execution. 

Given the UI log $\Sigma \in \MA^*$, we identify the multiset $W \in \mathcal{B}(\MA^*)$ of routine executions by segmenting $\Sigma$ at occurrences of actions from $\MA_F$. The log is split into subsequences $\mathcal{F}(\Sigma) = \{ t_1, t_2, \dots, t_p \}$ such that:
$t_j \in \MA^*, \quad \forall j \in [1, p], \quad \Sigma = \bigoplus_{k=1}^p t_k,
$
where the symbol $\bigoplus$ denotes sequence concatenation, and each $t_j$ contains exactly one action from $\MA_F$, appearing only as its final element. The resulting multiset of routine executions is as follows $
W = \biguplus_{t_j \in \mathcal{F}(\Sigma)} t_j
$

\paragraph{Encoding of the Routine Executions.}
\label{sec:encoding}
Following the extraction of the multiset of routine executions $W$, we transform all routine executions $[\lambda_1, \lambda_2, \ldots, \lambda_p] \in W$ into a multiset of feature vectors, which are subsequently clustered. 

Let $\langle a_1,\ldots,a_n \rangle$ be any ordering of the activities in $\MA$, namely $\MA=\cup_{i=1}^n a_i$. Each routine execution $\lambda \in W$ is encoded into a vector $(v_1,\ldots,v_n)$ where, for all $1 \leq i \leq n$, $v_i$ is equal to the number of executions of actions $a_i$ in $\lambda$.
For example, consider three routine executions: 
$
\lambda_1 = \langle a, b \rangle, \quad \lambda_2 = \langle a, c, c \rangle, \quad \lambda_3 = \langle b, c \rangle.
$
When encoded, their corresponding vectors are:  
$
\lambda_1 \rightarrow (1, 1, 0), \quad  
\lambda_2 \rightarrow (1, 0, 2), \quad  
\lambda_3 \rightarrow (0, 1, 1).
$

\paragraph{Clustering.}
\label{sec:clustering}

Once the different routine executions are encoded as vectors with dimensions related to the actions of the UI logs, these vectors and consequently the routine executions are clustered. Each cluster $C_i$ is associated with one routine type $T$ and becomes one routine log $C$. For each vector in $C_i$, the corresponding routine execution is added to $C_i$. 
The proposed technique is independent of the specific clustering method employed. However, we have equipped our implementation with three clustering methods: K-Means~\cite {hartigan1979algorithm}, DBSCAN~\cite{deng2020dbscan}, and HDBSCAN~\cite {stewart2022implementation}. We use the default distance metric, i.e., the Euclidean distance, provided by each algorithm in their standard configurations.  
\section{Experiments}
\label{sec:experiments}
To validate the effectiveness of our proposed technique, we conducted a series of experiments using the synthetic UI logs that were generated and employed by Leno et al.~\cite{DBLP:conf/icpm/LenoADRMP20}. The goal was to assess how well our technique discovers routine execution models compared to existing state-of-the-art techniques. These experiments focus on evaluating routine log extraction, conformance checking, and robustness to noise. The implementation and evaluation of the proposed technique, along with its comparison to state-of-the art techniques in Python, are available at \url{https://github.com/Faizanunipd/Routine-Model-Discovery-by-Extracting-Routine-Logs-in-RPA.git}. Section \ref{subsec1:experiment} discusses the data sets and evaluation methodology, and Section \ref{subsec2:experiment} provides a detailed analysis of the comparative results. 

\subsection{Data sets and Evaluation Methodology}
\label{subsec1:experiment}
For the evaluation, we used the nine UI logs that Leno et al.~\cite{DBLP:conf/icpm/LenoADRMP20} synthetically generated and employed in the experiments for their technique. These UI logs were chosen because they were provided with ground-truth models of the routines, which are supplied as Colour Petri nets in CPN-Tools format~\cite{bosco2019discovering}. These UI logs did not contain noise; a perfect clustering would guarantee that each routine execution in each log of any routine type $r$ is perfectly fitting with the ground-truth model of $r$. As indicated in Section~\ref{sec:intro}, we also aim to assess how our technique compares with those from the state of the art when humans are not consistent in executing instances of the same routine type (e.g., because of mistaken actions). This means that UI logs contain inconsistencies. This type of analysis of UI logs with inconsistencies was not considered in previous research works on routine-type discovery, neither when developing the techniques nor when assessing them. 

Algorithm~\ref{alg:noise} shows the pseudocode to add noise to a UI log $\Sigma$, so as to finally return a UI log $\Sigma'$. The algorithm requires a noise level $l \in [0,1]$, and builds $\Sigma'$ by subsequently concatenating events. The algorithm iterates over the action index $i \in [1,|\Sigma|]$: action $\Sigma(i)$ is concatenated to $\Sigma'$ with probability $(1-l)$, while it is removed with probability $0.5 \cdot l$ (lines 4-5). With the remaining probability of $0.5 \cdot l$, a new action is created for a random activity and concatenated to $\Sigma'$: however, the index $i$ is not increased, which means that $\Sigma(i)$ might still be concatenated at the next iteration cycle. 

\SetKwComment{Comment}{// }{ }

\SetKwFunction{Rand}{random}  
\SetKwFunction{CreateAction}{createAction}
\SetKwFunction{Append}{append}

\begin{algorithm*}[t!]
 
\caption{Noise Injection in a UI log segment}\label{alg:noise}
{

\KwIn{A UI log segment $\Sigma \in \MA^*$, Noise Level $l \in [0,1]$}
\KwOut{A noisy UI log segment $\Sigma' \in \MA^*$}
$\Sigma' \gets \langle \rangle$ \Comment*[r]{Initialize output log}
$i \gets 1$\;
\While{$i \leq |\Sigma|$}{
\tcc{$\Rand(0,1)$ generates a random number according to a uniform distribution in $[0,1]$}
    \If{$\Rand(0,1) \leq l$}{
        \If{$\Rand(0,1) \leq 0.5$}{
            $i \gets i + 1$ \Comment*[r]{Simulate missing action by skipping}
        }
        \Else{
            $\overline{e} \gets \CreateAction(\MA)$ \Comment*[r]{Create an action for a random activity in $\MA$ } 
            $\Sigma' \gets \Sigma' \bigoplus \langle \overline{e} \rangle$ \Comment*[r]{Concatenate random action}
        }
    }
    \Else{
        $e \gets \Sigma(i)$ \Comment*[r]{Access current action}
        $\Sigma' \gets \Sigma' \bigoplus \langle e \rangle$ \Comment*[r]{Concatenate current action to output}
        $i \gets i + 1$\;
    }
}
\Return $\Sigma'$\;}
\end{algorithm*}

The clustering obtained by our technique and by the state-of-the-art techniques has initially been assessed using the Jaccard coefficient (JC), as in Leno et al.~\cite{DBLP:conf/icpm/LenoADRMP20}. Specifically, given a set $\overline{\mathcal{C}}=\{ C_1,\ldots,C_n \}$ of routine logs, the metric computes the average over every routine $C_j \in \overline{\mathcal{C}}$ of
how well the set of actions in $C_j$ matches the most similar reference set $G_i$, which represents the set of actions in the $i^{th}$ ground-truth routine type:

\begin{definition}[Jaccard Coefficient]
\label{def:max-JC}
Let \(\overline{\mathcal{C}}=\{C_1, \ldots, C_n\} \subset \MB(\MA^*)\) be the set of routine logs discovered from a UI log. Let  $\MA_{C_i} = \bigcup_{a \in {C_i}} (a)$
be the set of actions in \(C_i\). Let \(\mathcal{G}=\{G_1,\ldots,G_n\}\) be the set of ground-truth actions for each of the \(n\) routine types, namely \(G_i \subseteq \MA\) for any \(1 \leq i \leq n\). The Jaccard Coefficient is computed as follows:
\begin{equation*} \label{eq:MaxJC}
\text{JC}(\overline{\mathcal{C}}, \mathcal{G}) = \avg_{C \in \overline{\mathcal{C}}} \Big(\max_{G_i \in \mathcal{G}} \frac{|\MA_{C} \cap G_i|}{|\MA_{C} \cup G_i|}\Big)
\end{equation*}
\end{definition}
However, Jaccard Coefficient (JC) only considers which activities are performed within routines, ignoring the order of the constituent actions. Since we know the exact model $m$ of each routine type, we can compute 
$LogModelFitness(C,m)$ between a routine log $C$ and a routine model $m$ using traditional process-mining fitness calculation~\cite{vanderAalst2016}. Then, adapting Definition~\ref{def:max-JC} to this context, we compute the average over all routine logs that contain traces (i.e., $|C| > 0$) of the maximum log-model fitness value with respect to each ground-truth model.
\begin{definition}[Fitness]\label{def:max_fitness}
Let $\overline{\mathcal{C}}=\{C_1, \ldots, C_n\} \subset \MB(\MA^*)$ be the set of routine logs discovered from a UI log.
Let \( C \) be a routine log. Let \( \mathcal{M} = \{ m_1, \dots, m_n \} \) be the set of ground-truth routine models. The fitness between the set of $\overline{C}$ of routine logs and the set of $\mathcal{M}$ models is defined as follows:

\begin{equation*} \label{eq:MaxFitness}
\text{Fitness}(\mathcal{\overline{\mathcal{C}}},  \mathcal{M}) = \avg_{C \in \overline{\mathcal{C}} \text{ s.t. } \mid C \mid >0} \Big(\max_{m_j \in  \mathcal{M}} \text{LogModelFitness}(C, m_j)\Big)
\end{equation*}
\end{definition}
Routine models for the nine UI logs are provided in the form of Petri nets, and we compute the alignment-based fitness~\cite{vanderAalst2016} of the routine logs against every routine-type model.  

For each UI log discussed in Section \ref{subsec1:experiment}, we extracted the set of routine logs, using our technique and other techniques from the literature, and compared the results in terms of Jaccard Coefficient (JC) and fitness. As mentioned above, we also introduced noise with levels $0.1, 0.2, 0.3$ and $0.4$ (cf.\ Algorithm~\ref{alg:noise}) to assess the robustness of the various techniques with increasing levels of noise. 
For each noise level larger than zero, noise was introduced 10 times per UI log. This iterative process mitigates bias from a single instance of noise and ensures a more reliable performance evaluation. 
The final results were obtained by averaging scores over multiple repetitions, accounting for variability due to stochastic factors in the noise injection during experimentation.

To illustrate the superiority of the results obtained by our technique, we also computed the Jaccard coefficients and the fitness values for three techniques from the state-of-the-art, namely by Leno et al.~\cite{DBLP:conf/icpm/LenoADRMP20}, Agostinelli et al.~\cite{DBLP:conf/icsoc/AgostinelliLM21}, and Rebmann et al.~\cite{DBLP:conf/caise/RebmannA23}. Leno et al.~\cite{DBLP:conf/icpm/LenoADRMP20} proposed four different variants based on length-based routine extraction, frequency-based routine extraction, cohesion-based routine extraction, and coverage-based routine extraction. 

\begin{figure*}[t!]
    \centering
    \begin{subfigure}{1\textwidth}
        \centering
        \includegraphics[width=\linewidth]{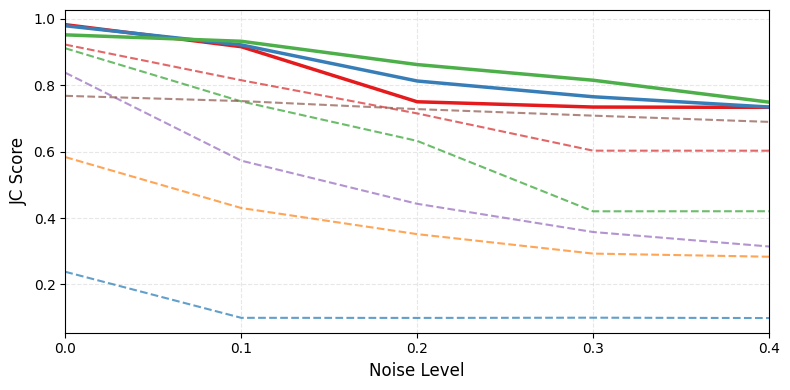}
        \caption{Jaccard Coefficient (JC)}
        \label{fig:JC}
    \end{subfigure}
    
    \centering
    \begin{subfigure}{1\textwidth}
        \centering
        \includegraphics[width=\linewidth]{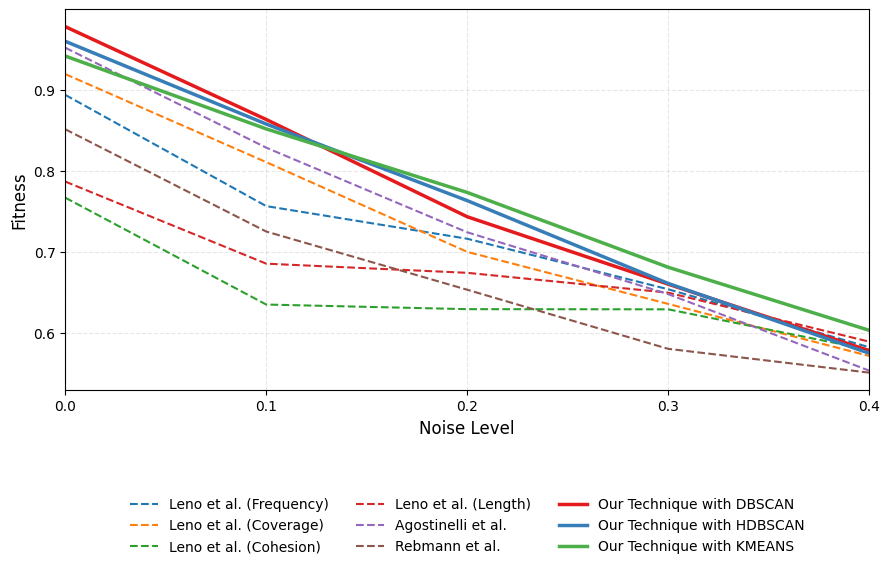}
     \caption{Fitness Score}
        \label{fig:fitnessNew}
    \end{subfigure}
    \caption{Trend of the JC and fitness values for the different techniques evaluated, under varying noise levels. The dotted lines represent state-of-the-art techniques, while the solid lines correspond to our techniques using different clustering methods.}
    \label{fig:jc_fitness}
\end{figure*}

Both Leno et al.~\cite{DBLP:conf/icpm/LenoADRMP20} and Agostinelli et al.~\cite{DBLP:conf/icsoc/AgostinelliLM21} only allow extracting the set of routine actions for each routine type, without extracting the routine logs. As mentioned, this would not enable the discovery of routine models. For the evaluation discussion in this section, it would not be possible to compute the fitness metrics in Definition~\ref{def:max_fitness}. With the goal in mind to extract the routine logs, each set $S_i$ of routine actions, which identifies a routine type $T_i$, is transformed into a vector $V_i$ of as many dimensions as the number of actions in the UI log. Each dimension is associated with an action, and takes the value 0 or 1 depending on whether or not the action is part of $S$. For example: Given a UI log with actions $\{A, B, C, D, E\}$, the action set for routine type $T_1$, denoted as \mbox{$S_1$ = $\{A, C, D\}$}, is represented by the binary vector $V_1 = (1, 0, 1, 1, 0)$. In this vector, each position corresponds to an action from the UI log, where $1$ indicates that the action is part of $V_i$ and $0$ indicates that it is not. Each routine execution is also converted into a vector, analogously to what is discussed in Section~\ref{sec:encoding}, and assigned to the routine type $T_i$ so that $V_i$ is the closest. 

\subsection{Evaluation Results}
\label{subsec2:experiment}
As discussed above, we evaluated the techniques from the state of the art by Leno et al.~\cite{DBLP:conf/icpm/LenoADRMP20}, Agostinelli et al.~\cite{DBLP:conf/icsoc/AgostinelliLM21}, and Rebmann et al.~\cite{DBLP:conf/caise/RebmannA23} and compared with our technique. For Leno et al.~\cite{DBLP:conf/icpm/LenoADRMP20}, we evaluated the four variants. Our technique was evaluated in three configurations where K-Means, DBSCAN and HDBSCAN were employed as clustering methods (cf.\ Section~\ref{sec:clustering}). 

The results of our experiments are reported in Figures~\ref{fig:JC} and~\ref{fig:fitnessNew}, in terms of values of JC and fitness, respectively. Each line refers to a different technique or variant. The x-axis indicates the noise level, and the y-axis refers to the JC or fitness value for every technique and noise level, namely 0 (i.e., no noise), 0.1, 0.2, 0.3 and 0.4. The points referring to each technique and noise level are the average over the nine UI logs employed in the experiments, where each experiment was run ten times for each log, technique and noise level.

The comparison of JC metric in Figure~\ref{fig:JC} shows that our technique achieves higher values of the metric, consistently
for all clustering techniques and noise levels.

The fitness computation excludes the routine logs that are generated without traces. In fact, this situation occurs when our technique is compared with Leno et al.~\cite{DBLP:conf/icpm/LenoADRMP20} and Agostinelli et al.~\cite{DBLP:conf/icsoc/AgostinelliLM21}. As a consequence, some routine types are assigned a set of actions that is a subset of those assigned to a different routine type. When routine executions are then assigned to routine types they are often associated to the routine type with the largest set of actions. This ultimately means that no routine associated with the routine types has a smaller set of actions. As indicated in Definition~\ref{def:max_fitness}, we exclude the empty routine logs, i.e., routine logs without traces, from the fitness computation because fitness cannot formally be computed on an empty log. 

Table~\ref{tab:empty_logs} illustrates the percentage of empty routine logs for each technique and noise level. It highlights a significant drawback of certain routine discovery techniques: the generation of empty routine logs. The method by  Agostinelli et al.~\cite{DBLP:conf/icsoc/AgostinelliLM21} consistently produces a high percentage of empty logs, over 75\% against all noise levels, indicating that many of its discovered routines fail to attract any traces. Similarly, the frequency and coverage variants of Leno et al. ~\cite{DBLP:conf/icpm/LenoADRMP20}  also yield substantial portions of empty logs, particularly under no noise conditions (e.g., 70\% and 51\% at noise level 0.0, respectively). In contrast, our technique and that by Rebmann et al.~\cite{DBLP:conf/caise/RebmannA23} do not produce empty routine logs, demonstrating stronger alignment between discovered routines and observed behavior.

\begin{table}[t!]
    \caption{Percentage of empty routine logs for different techniques and noise levels. 
    We do not report on Rebmann et al.~\cite{DBLP:conf/caise/RebmannA23} and our technique 
    because the empty-log percentage is always 0\%.}
    \label{tab:empty_logs}
    \scriptsize
    \vspace{0.5cm}
    \centering
    \renewcommand{\arraystretch}{1.2} 
    \begin{tabular}{|c|c|c|c|c|c|}
        \hline
        \multirow{2}{*}{\textbf{Noise Level}} & 
        \multicolumn{4}{c|}{\textbf{Leno et al. \cite{DBLP:conf/icpm/LenoADRMP20}}} &
        \multirow{2}{*}{\textbf{Agostinelli et al. \cite{DBLP:conf/icsoc/AgostinelliLM21}}} \\
        \cline{2-5}
         & Frequency & Coverage & Cohesion & Length & \\
        \hline
        0.0 & 70\% & 51\% & 13\% & 14\% & 89\% \\
        0.1 & 0\%  & 23\% & 25\% & 13\% & 79\% \\
        0.2 & 0\%  & 13\% & 12\% & 10\% & 79\% \\
        0.3 & 0\%  & 8\%  & 3\%  & 0\%  & 78\% \\
        0.4 & 0\%  & 5\%  & 3\%  & 0\%  & 76\% \\
        \hline
    \end{tabular}
\end{table}

Figure~\ref{fig:fitnessNew} shows that our technique generally achieves higher fitness scores than what state-of-the-art techniques do, for every noise level. While the K-Means variant performs slightly worse than Agostinelli et al. for noise level 0.0, both DBSCAN and HDBSCAN consistently outperform existing methods. However, the margin of improvement admittedly narrows progressively with increasingly higher noise levels. This is also partly due to the fact that, even in case of perfect creation of routine logs, the fitness decreases when the noise increases: for larger noise levels, there is a clear lower maximum in the fitness values that can be achieved, even in presence of a perfect assignment of routine executions to routine logs. 
Note that noise levels of 0.3 and 0.4 refer to UI logs with an extremely large variability of behavior of humans and a large extent of their mistakes. There is also an increased risk of misclassification. Specifically, when routines share overlapping actions, traces with higher levels of noise may be incorrectly clustered with a different routine type. This behavior is reflected in the observed decline in fitness at high noise levels. The analysis of the results in Figures~\ref{fig:JC} and~\ref{fig:fitnessNew} illustrates that, on average, our technique outperforms the state of the art for every noise level.

In conclusion, the findings confirm our initial hypothesis that the main benefit of our technique is to generate the routine logs, compared with the state-of-the-art techniques, which conversely only assign actions to routine types and do not generate routine logs. The results indicate that further extensions of state-of-the-art techniques, when applied as a post hoc approach to generating routine logs, do not reach the same level of quality in terms of Jaccard Coefficient (JC) and fitness metrics. 

Lastly, if we compare the performance of our technique for different clustering methods that are supported (cf.\ Section~\ref{sec:clustering}), K-Means achieved the highest performance. K-Means assigns all points to clusters, which leads to higher fitness and JC scores, especially under noisy conditions. HDBSCAN and DBSCAN perform better on cleaner data but degrade more under noise. Overall, K-Means is more robust to noise, while HDBSCAN and DBSCAN are preferable for clean, well-separated data.

\section{Experiments on Real-World UI Logs}
\label{sec:experiments2}
To further validate the robustness and generalizability of our proposed methodology, we extended our experiments to real-world UI log. Although the synthetic UI logs provided a controlled environment to assess the performance of our technique, real-world data introduces additional complexities, such as noise, human inconsistencies, and variability in behaviour. These factors are crucial to evaluating the true effectiveness of routine log extraction and clustering methods in practical scenarios. Section~\ref{subsec1:experiment2} presents the datasets description and preprocessing for filtering incomplete cases and section~\ref{sec:experiments3} presents the results of our technique on real-world UI log, comparing its performance against state-of-the-art methods, and highlighting the challenges and insights gained from this extended evaluation.

\subsection{Datasets: Description and Preprocessing}
\label{subsec1:experiment2}
For the evaluation on real-world UI log from a company that encompasses a large number of routine types. Unfortunately, several routine types have few executions recorded in the UI log. We thus decided to opt for those types that have at least a hundred routine executions. So we come up with the following routine types, namely sold to creation customer service, process 1, goal settings, ship to creation customer, supplier request buyer, impact review 2024, create catalog, add goals, create and assign team goal, create individual goal and home tour. Each routine execution is characterized by a special start event and a special end event. We observed that many executions did not have an end event and had a special action named close as the last event, which was filtered out before any further analysis. The close action means that the user did not complete the routine and ended the window by pressing or clicking the close icon.

Note that the original real-world UI log contained an actual routine-execution identifier, which was crucial to be able to find the incomplete routine executions. However, after the pre-processing step to remove those incomplete, we also removed the routine-execution identifier, because indeed our aim is to re-discover those executions from a UI log without identifiers. The routine identifiers were only used to determine the ground truth to compute the JC values and to discover routine models. The real-world UI log contains a total of 13,001 routine executions, which, after filtering, were reduced to 6,142 routine executions belonging to different routine types and used for further experimentation.

\subsection{Evaluation Results}
\label{sec:experiments3}


The experiments were conducted on the real-world datasets presented in section \ref{subsec1:experiment2}, and the performance of our technique was compared with that of the state-of-the-art methods by Leno et al.~\cite{DBLP:conf/icpm/LenoADRMP20}, Agostinelli et al.~\cite{DBLP:conf/icsoc/AgostinelliLM21} and Rebmann et al.~\cite{DBLP:conf/caise/RebmannA23} as discussed in section~\ref{sec:experiments}.

\begin{figure}[t!]
\centering
\includegraphics[width=\textwidth]{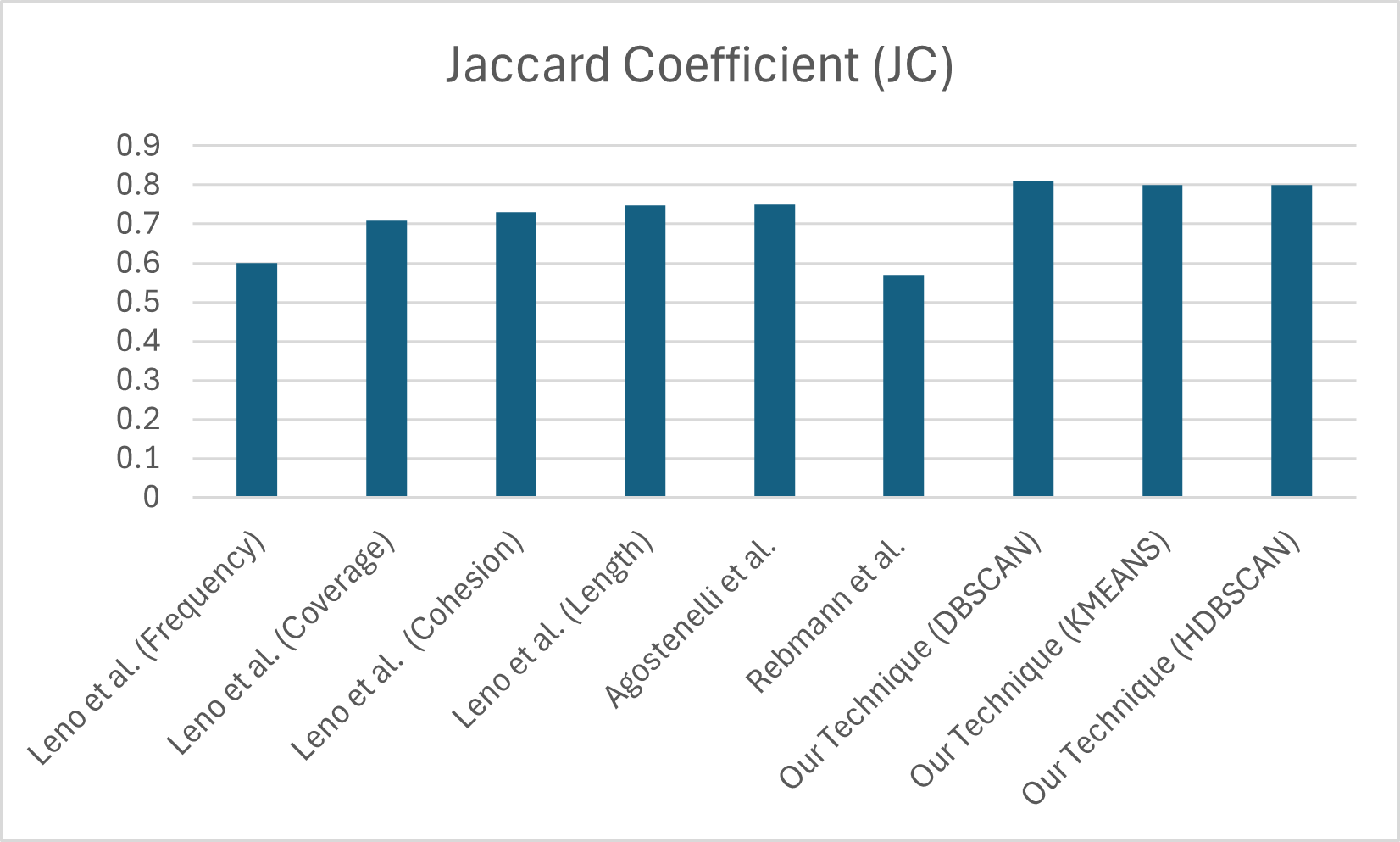}
\caption{Average Jaccard Coefficient (JC) values across all techniques. The x-axis lists the evaluated methods, while the y-axis shows the average JC score (0–1). Higher values indicate greater similarity between the extracted routines and the ground-truth routines.} 
\label{fig:JC_real}
\end{figure}

Figure~\ref{fig:JC_real} presents the bar plot of the Jaccard Coefficient (JC) values for the real-world UI log across all evaluated techniques. The x-axis lists the techniques, and the y-axis reports the JC score, which measures the similarity between the extracted routines and the ground-truth routines. Higher JC values (closer to 1) indicate stronger similarity. The results show that the baseline methods by Leno et al.~\cite{DBLP:conf/icpm/LenoADRMP20} and Agostinelli et al.~\cite{DBLP:conf/icsoc/AgostinelliLM21} achieve modest JC values and are frequently affected by empty logs. Rebmann et al.~\cite{DBLP:conf/caise/RebmannA23} obtain moderate JC scores, while our clustering-based approaches (DBSCAN, K-Means, HDBSCAN) reach higher values, indicating closer alignment with the ground truth.

As shown in Table~\ref{tab:empty_logs_in_real_dataset}, the generation of empty logs represents a major limitation of some techniques. For the real-world UI log, the Leno et al. (Frequency)~\cite{DBLP:conf/icpm/LenoADRMP20} method produces empty logs in 85\% of the cases, while the Leno et al. (Cohesion)~\cite{DBLP:conf/icpm/LenoADRMP20} produces 33\% empty logs. Agostinelli et al.~\cite{DBLP:conf/icsoc/AgostinelliLM21} also generate 63\% empty logs. These outputs cannot be considered valid representations of user behavior. In contrast, our clustering-based methods and Rebmann et al. never produce empty logs, ensuring that evaluation is performed on valid routine logs.

\begin{table}[t!]
    \caption{Percentage of empty routine logs for different techniques on the real-world UI log. 
    Techniques by Rebmann et al.~\cite{DBLP:conf/caise/RebmannA23} and our approach are omitted, 
    as they never produce empty logs (0\%).}
    \label{tab:empty_logs_in_real_dataset}
    \vspace{0.4cm}
    \centering
    \renewcommand{\arraystretch}{1.3} 
    \begin{tabular}{|c|c|c|c|c|c|}
        \hline
        \multirow{2}{*}{\textbf{Log}} & 
        \multicolumn{4}{c|}{\textbf{Leno et al.~\cite{DBLP:conf/icpm/LenoADRMP20}}} &
        \multirow{2}{*}{\textbf{Agostinelli et al.~\cite{DBLP:conf/icsoc/AgostinelliLM21}}} \\
        \cline{2-5}
         & Frequency & Coverage & Cohesion & Length & \\
        \hline
        Real-world UI log & 85\% & 0\% & 33\% & 0\% & 63\% \\
        \hline
    \end{tabular}
\end{table}

\begin{figure}[t!]
\centering
\includegraphics[width=\textwidth]{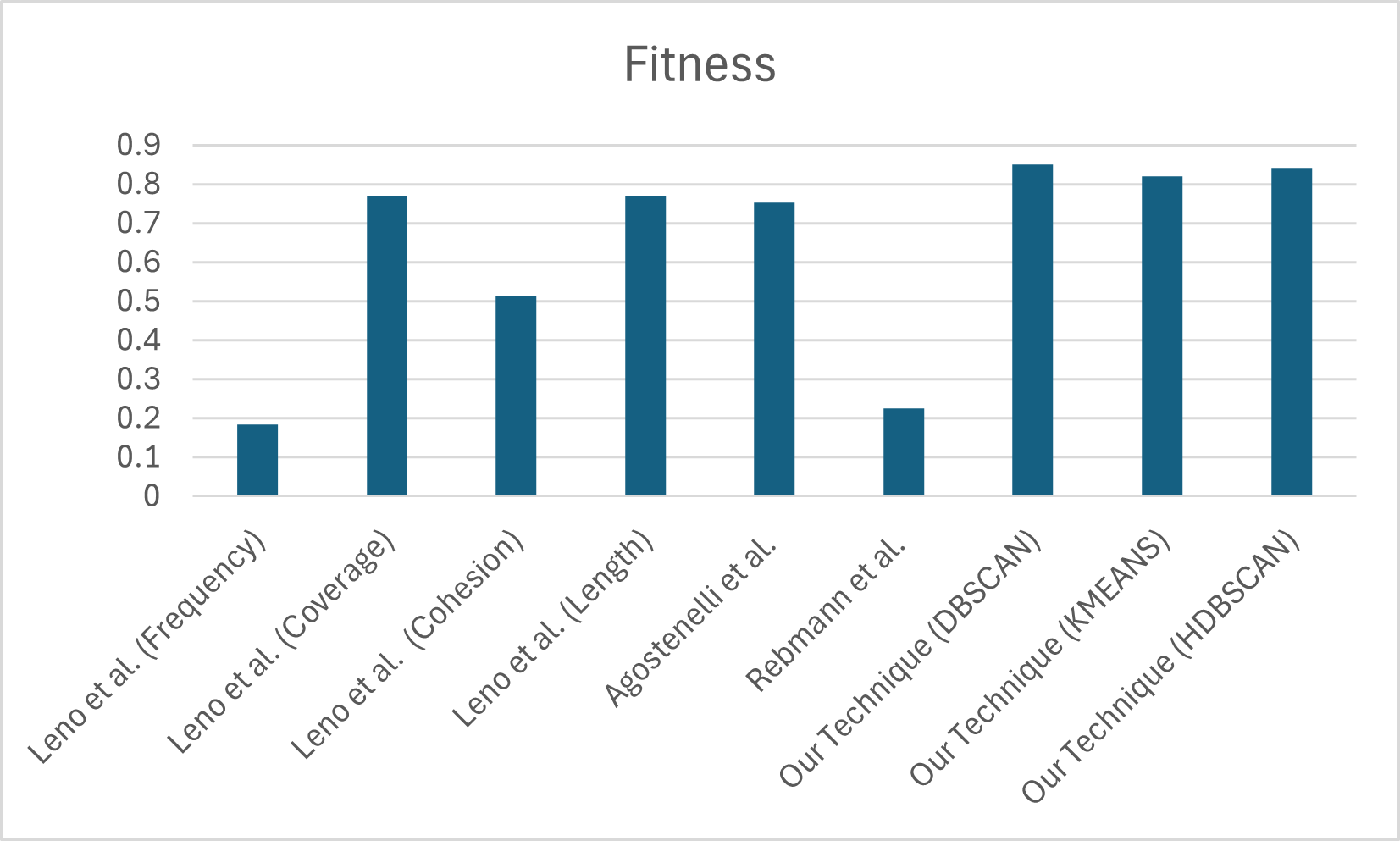}
\caption{Average Fitness values across all techniques. The x-axis lists the evaluated methods, while the y-axis shows the average fitness score (0–1). Higher values indicate stronger conformance of the discovered routine logs with the ground-truth models.} 
\label{fig:fitness_real}
\end{figure}

Figure~\ref{fig:fitness_real} presents the average Fitness values obtained by different techniques on the real-world UI log. The x-axis lists the evaluated methods and the y-axis reports the fitness scores, which quantify the degree to which the reconstructed routine logs reproduce the behavior of the ground-truth models. Higher values indicate stronger conformance, while lower values reflect weaker alignment. The results show that the Leno et al. (Coverage)~\cite{DBLP:conf/icpm/LenoADRMP20} and Leno et al. (Length)~\cite{DBLP:conf/icpm/LenoADRMP20} achieve moderate to high fitness, while Frequency and Cohesion yield considerably lower values. Agostinelli et al.~\cite{DBLP:conf/icsoc/AgostinelliLM21} produce modest fitness results, consistent with the presence of empty logs. Rebmann et al.~\cite{DBLP:conf/caise/RebmannA23} achieve moderate fitness values, while our clustering-based approaches obtain consistently higher values, approaching near-perfect alignment with the ground truth.

The results in Figures~\ref{fig:JC_real} and~\ref{fig:fitness_real} show the average JC and Fitness values for the real-world UI log across all the techniques evaluated. Our technique (DBSCAN, K-Means, and HDBSCAN) generate routine logs that align with the actual execution, while Rebmann et al.~\cite{DBLP:conf/caise/RebmannA23} consistently produce a single routine containing all executions. This avoids empty logs but reduces the variability of the discovered behavior. In contrast, Agostinelli et al. and the variants of Leno et al. are often affected by empty routine logs, which limits their ability to represent user behavior.

In summary, the analysis highlights key differences among existing techniques: some methods may oversimplify all executions into a single routine, others may produce empty logs, while our technique consistently provide more accurate routine logs. These findings demonstrate the importance of approaches that are capable of handling the complexities of real-world Robotic Process Mining (RPM) applications, where the accuracy of routine logs is essential.

\section{Conclusion}
\label{sec:conclusion}
This paper starts from the belief that RPA require models of the types of routines that are targeted. RPA tools indeed requires a formal model of how the routines should be executed. Although these models can be designed manually by experts, the risk is high that these hand-made models reflect the experts' perceptions, which might not accurately match how the routines are actually performed. It is thus imperative to try to discover the models of the different routine types from the so-called UI logs, which record how human actors perform the routine actions via information systems and software. As UI logs contain executions of different routine types, we need to split them into distinct routine logs, one per routine type, which can then be used as input for model discovery of each routine type. This is far from being trivial: we do not know in advance the actions that belong to each routine type, and the same action can also be shared among multiple routine types. 

Current literature predominantly focuses on techniques that use UI logs to solely assign human actions to routine types, but neglects the creation of routine logs, which are an essential input for discovering routine-type models. Moreover, existing approaches overlook the fact that humans often act inconsistently or make errors, referred to as noise,
such as copying the wrong cells in a spreadsheet and correcting them through additional actions. If techniques for extracting routine logs are not robust to noise, the resulting routine logs may be flawed, which lead to inaccurate and imprecise routine-type models when used as input to discovery techniques. 

The above consideration motivated us to propose a technique
that paves the way for improved routine-type model discovery that enhances process discovery in RPM. 
In particular, the technique is able to extract routine logs from UI logs, even in the presence of noise. Our technique leverages cluster methods while being independent of the peculiarities of any of them, and its implementation supports K-Means, DBSCAN, and HDBSCAN.

Extensive experimentation was conducted on nine UI logs with varying levels of noise induced up to 0.4 using a bootstrap technique. The assessment was based on state-of-the-art evaluation metrics, namely the JC and fitness score. The results showed that our technique produces more accurate routine logs, and is consistently more resilient to noise. Our experimental results also indicate that the technique is computationally efficient across UI logs of varying sizes, demonstrating practical scalability.

We conclude by highlighting some threat of validity and directions for future work. First, our technique requires a user-defined set of completion actions to perform segmentation: While this assumption typically holds in practice (e.g., routines terminate after pressing a button, saving a file, sending an email), we acknowledge that this requires the provision of domain knowledge, which might sometimes not be present. 



Importantly, the assessment was conducted using synthetic data, the same data sets employed by Leno et al.~\cite{DBLP:conf/icpm/LenoADRMP20} in their evaluation. This choice is motivated by the need for ground-truth models of the different routine types, which are essential for calculating the fitness and JC scores. While evaluating with real-world UI logs would be a valuable complementary analysis, it would not offer a concrete basis for assessing the quality of routine log identification, neither for our approach nor for existing methods in the literature.

\bibliographystyle{splncs04}
\bibliography{biblio}

\end{document}